\documentclass[journal]{IEEEtran}
\usepackage{blindtext}
\usepackage{graphicx}
\usepackage{hyperref}
\usepackage{dcolumn}
\usepackage{bm}
\usepackage{amsmath}
\usepackage{mathtools}
\usepackage{colonequals}
\usepackage{ulem}
\usepackage{amssymb}

\usepackage{algorithm}
\usepackage[noend]{algpseudocode}
\ifCLASSINFOpdf
\else
\fi
\begin{document}
%
\title{Entropic Determinants of Massive Matrices}
%
%
%

\author{Diego~Granziol and~Stephen~Roberts
\thanks{Machine Learning Research Group, University of Oxford}
}

\maketitle

\begin{abstract}

The ability of many powerful machine learning algorithms to deal with large data sets without compromise is often hampered by computationally expensive linear algebra tasks, of which calculating the log determinant is a canonical example. In this paper we demonstrate the optimality of Maximum Entropy methods in approximating such calculations. We prove the equivalence between mean value constraints and sample expectations in the big data limit, that Covariance matrix eigenvalue distributions can be completely defined by moment information and that the reduction of the self entropy of a maximum entropy proposal distribution, achieved by adding more moments reduces the KL divergence between the proposal and true eigenvalue distribution. We empirically verify our results on a variety of SparseSuite matrices and establish best practices.

\end{abstract}

\begin{IEEEkeywords}
Maximum entropy methods, approximation methods, Matrix Theory, constrained optimization, noisy constraints, log determinants.
\end{IEEEkeywords}

%
\IEEEpeerreviewmaketitle
\section{Motivation}
Scalability is one of the key challenges facing machine learning algorithms. In the era of large data sets, inference schemes are required to deliver optimal results within a constrained computational cost.
Linear algebraic operations with high computational complexity pose a significant bottleneck to algorithmic scalability, and the log determinant of a matrix~\cite{Bai1997} falls firmly within this category of operations.
The typical solution, involving Cholesky decomposition~\cite{Golub1996} for a general $n\times n$ positive definite matrix, $A$, entails time complexity of $\mathcal{O}(n^{3})$ and storage requirements of $\mathcal{O}(n^{2})$, which is unfeasible for large matrices. We further find that, along with making multiple matrix copies, typical implementations of Cholesky decomposition require contiguous memory. Consequently, the difficulty in calculating this term greatly hinders widespread use of the learning models where it appears, which includes determinantal point processes~\cite{Macchi1975}, Gaussian processes~\cite{Rasmussen2006}, and graph problems~\cite{Wainwright2006}.

\section{Contributions of this Paper}
\begin{figure}[t]
	\centering 
	\includegraphics[width=0.5\textwidth]{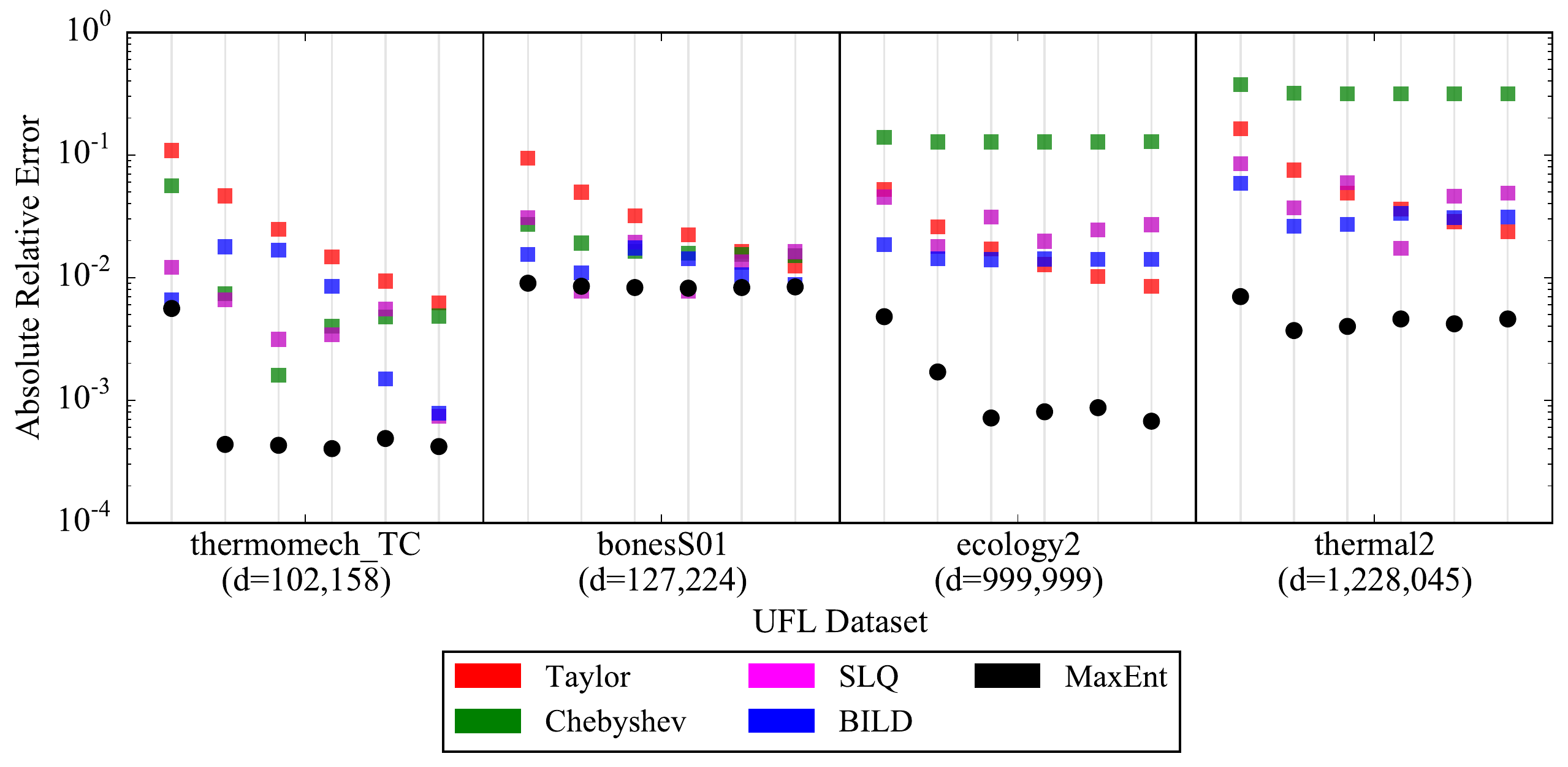}
	\caption{Absolute relative error of log determinant calculations on a collection SuiteSparse datasets using Stochastic trace estimate input data. MaxEnt (black dots) substantially outperforms other methods, figure originally from \cite{ete}}
	\label{fig:UFLcomparison}
\end{figure}

Recent work combining Maximum Entropy algorithms with stochastic trace estimates of moments displayed state of the art performance on log determinant estimates with an $\mathcal{O}(n^{2})$ computational time on both randomly generated and sparse matrices \cite{ete}, with results shown in Figure \ref{fig:UFLcomparison}. In this paper we address and answer many open pedagogical and practical concerns, such as:
\begin{enumerate}
\item Why should we characterize an eigenvalue probability distribution by its moments? To what extent do they embody relevant information?
\item What is the equivalence between sample averages and mean value constraints? When are they identical?
\item Can we characterize an eigenvalue distribution better with more moment constraints? Why do the Maxent predictions in Figure \ref{fig:UFLcomparison} from \cite{ete} get worse beyond a certain number of included moments?
\item If a practitioner wants to use MaxEnt algorithms and stochastic trace estimates to generate an accurate log determinant estimate of a large matrix, how many samples and how many moments do they need to take?
\end{enumerate}

\section{Can moments fully describe probability distributions?}
\label{momentsanddistributions}
For a probability measure $\mu$ having finite moments of all orders $\alpha_{k} = \int_{-\infty}^{\infty} x^{k}\mu(dx)$, if the power series $\sum_{k}\alpha_{k}/k!$ has a positive radius of convergence, that $\mu$ is the only probability measure with the moments $\alpha_{1},\alpha_{2},...$ \cite{billingsley_2012}. The proof essentially shows that such measures must share a characteristic function, which by the uniqueness theorem for characteristic functions, implies a unique measure and rests on the result that as $n \rightarrow \infty$ the ratio of the $n^{th}$ absolute moment to $n!$ goes to $0$ i.e.
\begin{equation}
\label{conditionconvergencemoments}
\frac{\int |x|^{n}\mu(dx)}{n!} \xrightarrow{n\rightarrow\infty} 0.
\end{equation}
Which means that in the $n \rightarrow \infty$ limit the growth of absolute moments can be at most n, namely,
\begin{equation}
\frac{\beta_{n}}{\beta_{n-1}} \xrightarrow{n\rightarrow \infty} \leq n.
\end{equation}
\subsection{Application to Entropic Trace Estimation}
Consider a random variable $z$ with mean $m$ and variance $\sigma$ using the property of the expectations of quadratic forms, the expectation 
\begin{equation}
\mathbb{E}[zz^{t}] = \sigma + mm^{t} = I,
\end{equation}
where we have assumed that the variable is zero-mean and unit-variance. This allows us to calculate the trace of any matrix power $A^{m}$ as
\begin{equation}
\begin{aligned}
& Tr(A^{m}) = Tr(A^{m}I) = Tr(A^{m}\mathbb{E}[zz^{t}]\\
& = \mathbb{E}[Tr(A^{m}zz^{t})] = \mathbb{E}[z^{t}A^{m}z].
\end{aligned}
\end{equation}
To prove that \eqref{conditionconvergencemoments} holds for all Covariance matrices. We note from linear algebra that an $N \times N$ matrix $A$ is diagonalizable if and only if it has $n$ linearly independent eigenvectors. All normal matrices, of which real-symmetric (covariance) matrices are a subset, are diagonalizeable. Hence the eigenvectors of $A$ span the space of the $\mathbb{R}^{n}$. We can thus write any vector as a linear combinations of the eigenvectors of $A$, $z = \sum_{i}\alpha_{i}|\phi_{i}\rangle$, where we have used dirac bra-ket notation to avoid confusion between the scalar $\alpha_{i}$ and the (ket) vector $|\phi_{i}\rangle$ the conjugate transpose (in this case just transpose as we are in the real space of ket is denoted as $z^{t} = \sum_{i}\alpha_{i}\langle \phi_{i}|$. Hence,
\begin{equation}
\label{proofeteworks}
\begin{aligned}
&\mathbb{E}(z^{t}A^{n}z) = \mathbb{E}\bigg(\sum_{j}\alpha_{j}\langle \phi_{j}| A^{n} \sum_{i}\alpha_{i}|\phi_{i}\rangle\bigg)\\
&=\mathbb{E}\bigg(\sum_{i,j}\alpha_{j}\alpha_{i} \lambda^{n} \langle \phi_{j}|\phi_{i}\rangle\bigg) = \mathbb{E}\bigg(\sum_{i}|\alpha_{i}|^{2} \lambda^{n} \bigg)\\
&=\mathbb{E}\bigg(\lambda_{max}^{n}|\alpha_{max}|^{2}\bigg[1 + \sum_{i\neq imax}\bigg|\frac{\alpha_{i}}{\alpha_{max}}\bigg|^{2}\bigg(\frac{\lambda_{i}}{\lambda_{max}}\bigg)^{n}\bigg]\bigg)\\
&\xrightarrow{n\rightarrow\infty}\mathbb{E}\bigg(\lambda_{max}^{n}|\alpha_{max}|^{2}\bigg) = \mathbb{E}\bigg(|\alpha_{max}|^{2}\bigg)\lambda_{max}^{n}.
\end{aligned}
\end{equation}
In which we have used ortho-normality $\langle \phi_{i}|\phi_{j}\rangle = \delta_{i,j}$ along with the fact there are $n$ distinct eigenvalues and that $\lambda_{i\neq\max} < \lambda_{max}$. Although this is not strictly necessary, if there were multiple degenerate maximum eigenvalues, the eigenvalue pre-factor in \eqref{proofeteworks} would become:
\begin{equation}
\mathbb{E}\bigg(|\alpha_{max}|^{2}\bigg) \xrightarrow{\text{k degenerate maxima}} \mathbb{E}\bigg(\sum_{i}^{k}|\alpha_{i}|^{2}\bigg).
\end{equation}
Given that the $n^{th}$ raw moment of the eigenvalue spectrum can be written as $\mathbb{E}(\lambda^{n}) = (1/n)Tr(A^{n})$ we can thus relate equation \eqref{conditionconvergencemoments} to equation \eqref{proofeteworks},
\begin{equation}
\begin{aligned}
& \frac{\mathbb{E}(\lambda^{n})}{n!} = \frac{(1/n)Tr(A^{n})}{n!} = \mathcal{C}\\
& \log \mathcal{C} = n\log \lambda_{max} + \log \mathbb{E}(|\alpha_{max}|^{2})\\
& - (n+1)\log(n+1) + (n+1)\\
& \log \mathcal{C} \xrightarrow{n \rightarrow \infty} n(\log \lambda_{max} - \log n +1) \xrightarrow{\forall \lambda_{max}} -\infty \\
& C \xrightarrow{n \rightarrow \infty} 0\\
& QED\\
\end{aligned}
\end{equation}
Here we use the fact that, for positive semi-definite matrices, all moments are positive and hence raw moments are equivalent to absolute power moments along with Stirling's approximation in the large $n$ limit. This proves that it is possible to uniquely define a Covariance matrix's eigenvalue probability distribution through its moment information. 

This answers the question as to why beyond being computationally cheap $\mathcal{O}(n^{2})$, it is worth sampling moments. They embody relevant information. 

\section{Maximum Entropy}
The method of maximum entropy (MaxEnt) \cite{maxentreview} is a method which generates the least biased estimate of a proposal probability distribution, $q(x)$, given information in the form of functional expectations (also known as constraints). It is maximally non-committal in regards to missing information \cite{inftheoryjaynes}. Mathematically we maximize the functional,
	\begin{equation}
	\label{BSG}
	S = \int p(\vec{x})\log p(\vec{x})d\vec{x}- \sum_{i}\lambda_{i}\bigg[\int p(\vec{x})f_{i}(\vec{x})d\vec{x} - \mu_{i}\bigg],
	\end{equation}
with respect to $p(\vec{x})$, where $\langle f_{i}(\vec{x})\rangle = \mu_{i}$ are the values of the imposed mean value constraints. For stochastic trace estimation, the functions are the power moments, $f_{i} = x^{i}$. The first term in Equation \eqref{BSG} is the Boltzmann-Shannon-Gibbs (BSG) entropy. This has been applied in a variety of disparate fields, from modelling crystal defects in lattice models in condensed matter physics \cite{giffin} to inferring asset price movement distributions from option prices in finance \cite{nerioptions,entropybuchen}. It can be used to derive statistical mechanics (without the a priori assumptions of ergodicity and metric transitivity \cite{diego}), non-relativistic quantum mechanics, Newton's laws and Bayes' rule \cite{Gonzalez2014,caticha2012entropic}. It can be proved under the axioms of consistency, uniqueness, coordinate invariance, subset and system independence, that for mean value constraints any self consistent inference scheme must either maximize the entropic functional \eqref{BSG}, or any functional sharing its maximum \cite{shore1980axiomatic,maxentreview}. The Johnson and Shore axioms state that the entropy must have a unique maximum \cite{shore1980axiomatic} and, given the convexity of the BSG entropy, it contains a unique maximum provided that the constraints are convex. This is satisfied for any polynomial in $x$ and hence entropy maximization, given moment information, constitutes a self consistent inference scheme \cite{maxentreview}.

\section{Functional Expectations}
We provide a mathematical justification for the asymptotic equivalence between mean value constraints and sample expectations. 

From Chebyshev's inequality we have,
\begin{equation}
P(|X|\geq a) \leq \frac{1}{a^{p}}E(|X|^{p}),
\end{equation}
which, when applied to a set of independent random variables possessing a mean and variance, leads to:
\begin{equation}
\begin{aligned}
& P\bigg( \bigg|\frac{X_{1}+X_{2}+..+X_{N}}{n}-\mu \bigg| \geq \epsilon \bigg) \leq  \frac{1}{\epsilon^{2}}\frac{\sum_{i=1}^{n}Var(X_{i})}{n^{2}}\\  &\xrightarrow{i.i.d} \frac{1}{\epsilon^{2}}\frac{\sigma^{2}}{n} \xrightarrow{n\rightarrow\infty} 0.
\end{aligned}
\end{equation}
Here we have used the Chebyshev inequality and the fact that the variance of a sum is the sum of the variances (for independent random variables), followed by the i.i.d. assumption and the asymptotic limit respectively. This is known as the weak law of large numbers, as the limit is outside the brackets. Note that this limit does not necessitate the variables to be identically distributed, nor does it preclude weak dependence.\footnote{For fully dependent variables $Var(\sum_{i}^{n}X_{i}) = n^{2}Var(X)$ and hence the limit is never reached.} Using extensions of the central limit theorem \cite{stein1972,convergenceclt} it can be shown that for at most weakly dependent random variables obeying weak conditions, this result also holds.

As functions of random variables are themselves random variables, we can apply the same limit, i.e.
\begin{equation}
P\bigg( \bigg|\frac{F_{m}(X_{1})+..F_{m}(X_{N})}{n}-\langle F_{m}(X)\rangle \bigg| \geq \epsilon \bigg)  \xrightarrow{n\rightarrow\infty} 0.
\end{equation}
Considering each $F_{m}$ to be a stochastic trace estimate of the power moment $\int p(x)x^{m}dx$, we see that in the large $n$ or big data limit, we recover the true mean value constraint with probability $1$. 

\subsection{Sufficiency of Statistics}
All distributions derived from the method of maximum entropy are within the exponential family, i.e. they are of the form,
\begin{equation}
p(x|\theta) = A(x)\exp(\langle T(x),\theta\rangle - F(\theta)),
\end{equation}
where $\theta \in \Theta \subset \mathbb{R}^{d}$. $T$ and $A$ are fixed functions that characterize the exponential family, $F(\theta)$ is a normalization factor with respect to some measure $\nu(x)$ and $A(x)$ is the carrier measure. For variables independently drawn from the probability measure, we have
\begin{equation}
\begin{aligned}
&p(x_{1},x_{2}.....x_{n}|\theta) = \Pi_{i}p(x_{i}|\theta) \\
& = A(x)\exp(\langle \sum_{i} T(x_{i}),\theta\rangle - F(\theta)).
\end{aligned}
\end{equation}
This distribution hence depends on the input data only through the sample statistic $\sum_{i}T(x_{i})$, referred to as a sufficient statistic. Given the nature of our proposed inference scheme (MaxEnt) restricts us to the exponential family, it makes sense for us to compress the data with no loss of information. That no information is lost is implicit in the definition of a sufficient statistic and can be demonstrated using the data-processing inequality \cite{cover2012elements}. This insight is also discussed by Jaynes \cite{jaynes1982rationale}.

\section{self entropy as a divergence}
Consider the KL divergence $\mathcal{D}_{kl}$, also known as the minimum discrimination information, or negative relative entropy \cite{cover2012elements} between a true eigenvalue distribution $p(x)$ and a proposal MaxEnt solution $q(x) = \exp(\sum_{j}\alpha_{j}x^{i})$:
\begin{equation}
\label{kldiv}
\mathcal{D}(P||Q) = \int p(x)\log p(x)dx - \int p(x)\log q(x) 
\end{equation} note that the (self) entropy of the MaxEnt solution is given by
\begin{equation}
\begin{aligned}
\label{ent1}
& \mathcal{S}(Q) = -\int q(x)\log q(x) dx \\
&= \sum_{i}\alpha_{i}\int x^{i}\exp\bigg(-\sum_{j}\alpha_{j}x^{j}\bigg)dx = \sum_{i}\alpha_{i}\langle x^{i} \rangle,
\end{aligned}
\end{equation}
where $\alpha$ denotes the Lagrange multipliers pertaining to the MaxEnt solution and $\langle x^{j} \rangle$ refers to the expectation of the $\text{j}^{th}$ moment. 

The first term in equation \eqref{kldiv} is the negative entropy of the true unknown distribution $\mathcal{S}(p)$. We can thus rewrite equation \eqref{kldiv} as:
\begin{equation}
\label{sumentropies}
\begin{aligned}
&-\mathcal{S}(P) + \int p(x)\sum_{i}\alpha_{i}x^{i} = -\mathcal{S}(P) + \sum_{i}\alpha_{i}\langle x^{i} \rangle \\
& = -\mathcal{S}(P) + \mathcal{S}(Q).
\end{aligned}
\end{equation} 
We have used the fact that the functional expectations of our MaxEnt distribution by construction (Equation \eqref{BSG}) match that of the underlying distribution.  

Thus by minimizing $\mathcal{S}(Q)$, for which we have an analytic form, we manifestly reduce the KL divergence between our MaxEnt proposal $q(x)$ and our true eigenvalue distribution $p(x)$. It is further clear by the use of the information inequality \cite{cover2012elements} that the entropy of our proxy MaxEnt solution serves as an upper bound to that of the true solution, i.e.
\begin{equation}
\label{dataprocessesing}
\mathcal{D}_{kl}(P||Q) = \mathcal{S}(Q) -\mathcal{S}(P) \geq 0 \rightarrow \mathcal{S}(Q) \geq \mathcal{S}(P).
\end{equation}

\subsection{Consequences}
\label{epsilonkl}
In section \ref{momentsanddistributions} we proved that an eigenvalue distribution could be completely specified by its power moments. In the above section we show that that the entropy of the MaxEnt proposal distribution $q(x)$ is an upper bound to the entropy of the true data generating distribution $p(x)$ and prove that the reduction in self entropy $\mathcal{S}(q)$ is equivalent to reducing $\mathcal{D}_{kl}(p||q)$. In the next section we prove that adding information in the form of extra moment information necessarily reduces the self entropy $\mathcal{S}(q)$. This result generates an active procedure in which we can be principled in knowing how many functional expectations we need to take. We just sequentially calculate the self entropy of proposal MaxEnt distribution $q(x)$ using equation \eqref{ent1} and terminate the procedure at the point at which this decrease becomes negligible $-\Delta \mathcal{S} < \epsilon$.

\section{Lagrangian Duality}
Consider a generic optimization problem of the form,
\begin{equation}
\begin{aligned}
&\text{minimize } \thinspace f_{0}(x)\\
&\text{subject to } \thinspace f_{i}(x) \leq 0, \thinspace i=1...m\\
&\text{subject to } \thinspace h_{i}(x) = 0, \thinspace i=1...p
\end{aligned}
\end{equation}
where $x \in \mathbb{R}^{n}$ and the domain $\mathcal{D} = \bigcap\limits_{i=0}^{m} f_{i}\cap\bigcap\limits_{i=1}^{p}h_{i}$. We define the Lagrangian dual function as the infimum of the Lagrangian over the domain of $x$,
\begin{equation}
\begin{aligned}
& g(\lambda,\nu) = \underset{x \in \mathcal{D}}{inf}L(x,\lambda, \nu)\\
& = \underset{x \in \mathcal{D}}{inf}\bigg( f_{0}(x) + \sum_{i=1}^{m}\lambda_{i}f_{i}(x) + \sum_{i=1}^{p}\nu_{i}h_{i}(x)\bigg).
\end{aligned}
\end{equation}
As the dual is the pointwise infimum of a family of affine functions of $(\lambda, \nu)$, it is concave, irrespective of the convexity of $f_{0},f_{i},h_{i}$. \cite{boyd_vandenberghe_2009}. It is easily verifiable due to the net negativity of the two summation terms in $g(\lambda, \nu)$ that the dual provides a lower bound on the optimal value $p^{*}$ of the primal problem. This is known as weak duality. In the case of equality constraints this bound is tight. 

For general inequality constraints the difference between the primal and dual optimal solution (duality gap) is not 0. However, for $f_{0}...f_{m}$ convex, Affine equality constraints and certain regularity conditions, we have a duality gap of $0$, this is known as strong duality. An example of such a constraint qualification is Slater's condition, which states that there is an $x \in \textbf{relint } \mathcal{D}$ which satisfies the constraints.

\subsection{Application to Probability Distributions}
We consider a probability distribution $p:\mathcal{R}^{n} \rightarrow \mathcal{R}$ which satisfies the general axioms of non-negativity, associativity and normalizability. This defines a very general space of probability theories, of which the Bayesian is a special case \cite{walley_1991}. Thus $p(x) \geq 0$ for all $x \in C$ and $\int p(x)dx = 1$, where $C \subseteq \mathcal{R}^{n}$ is convex. The last condition follows from the definition of convexity and the fact that any sum of two real numbers is a real number. Then as any non negative weighting of a convex set preserves convexity,
\begin{equation}
\int_{C}p(x)x\thinspace dx \in C \thinspace
\end{equation}
if the integral exists.

\subsection{Application to Maximum Entropy}
\label{entropydecreaseproof}
We wish to maximise the entropic functional $\mathcal{S}(p) = -\int p(x)\log p(x)dx$ under certain moment constraints $\int p(x)x^{m}dx = \mu_{m}$. This can be written as,
\begin{equation}
\begin{aligned}
&\text{minimize } \thinspace f_{0}[p(x)] = \int p(x)\log p(x)dx\\
&\text{subject to } \thinspace h_{i}[p(x)] = \int p(x)x^{i}dx - \mu_{i} = 0, \thinspace i=1...p.
\end{aligned}
\end{equation}
Given that the negative entropy is a convex objective and that the moment equality constraints are affine in the variable being optimised over $p(x)$ by strong duality we have an equivalence between the solution of the dual and that of the primal.

It is also clear that the domain defined as the intersection of the constraint sets can never increase upon the addition of an extra constraint. Hence,
\begin{equation}
\underset{x \in \mathcal{D}=\bigcap\limits_{i=0}^{m}f_{i}}{inf}L(x,\lambda, \nu) \leq \underset{x \in \mathcal{D}=\bigcap\limits_{i=0}^{m+1}f_{i}}{inf}L(x,\lambda, \nu)
\end{equation}
and thus the entropy can only decrease when adding an extra constraint. Hence by adding more moment constraints, we always reduce the entropy and given equations \eqref{dataprocessesing} and \eqref{sumentropies} we necessarily reduce $\mathcal{D}_{kl}(p[x]||q[x])$, where $p[x], q[x]$ define the true eigenvalue and MaxEnt proposal distributions respectively.

\section{Noise}
\begin{figure}[t]
	\centering 
	\includegraphics[width=0.5\textwidth]{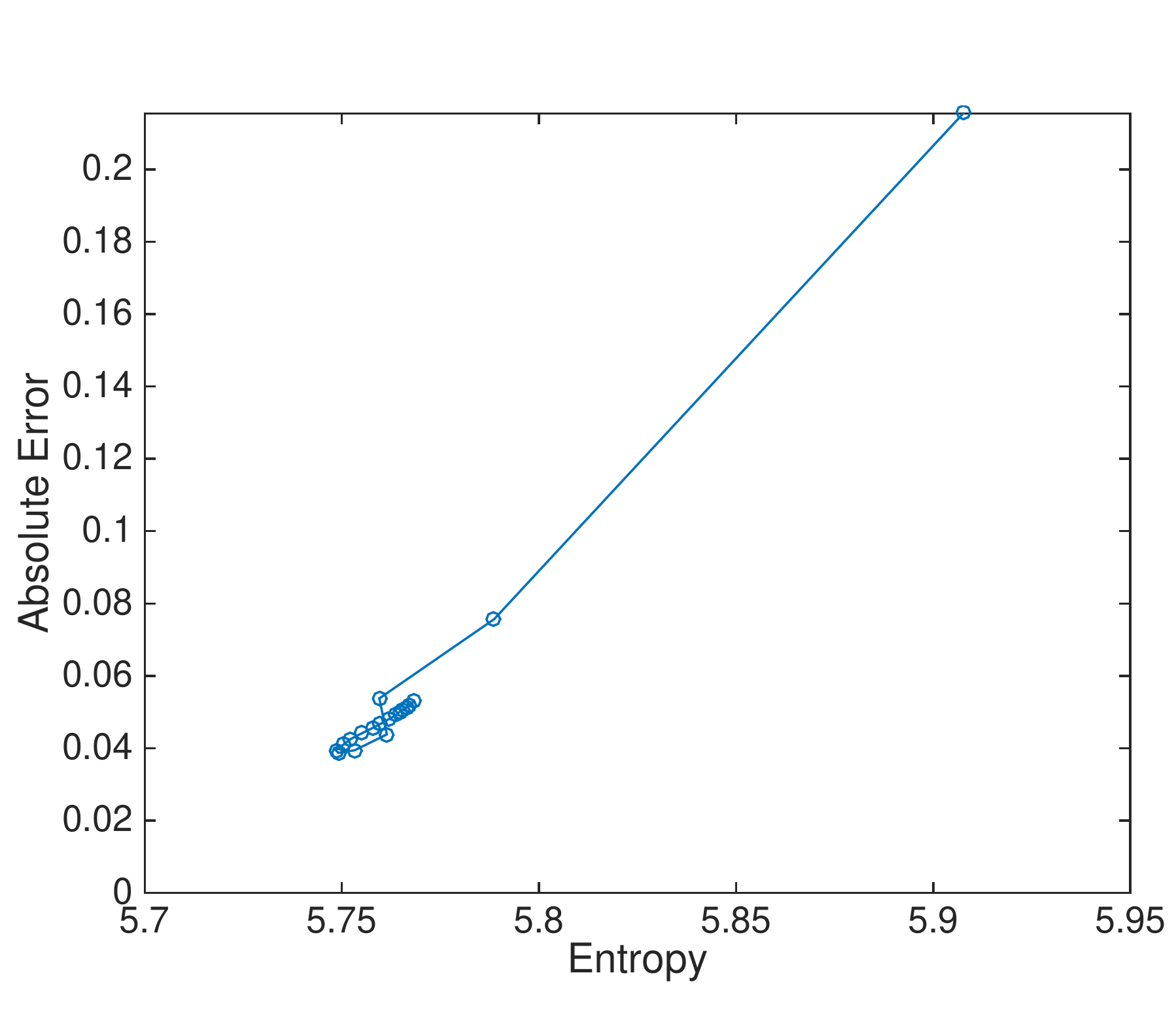}
	\caption{Plot of Absolute Error vs Entropy for Thermomech TC SuiteSparse dataset on 5 stochastic trace estimate samples}
	\label{fig:cropped_thermoent5sampleserrorentropy-croppedy}
\end{figure}

\begin{figure}[t]
	\centering 
	\includegraphics[width=0.5\textwidth]{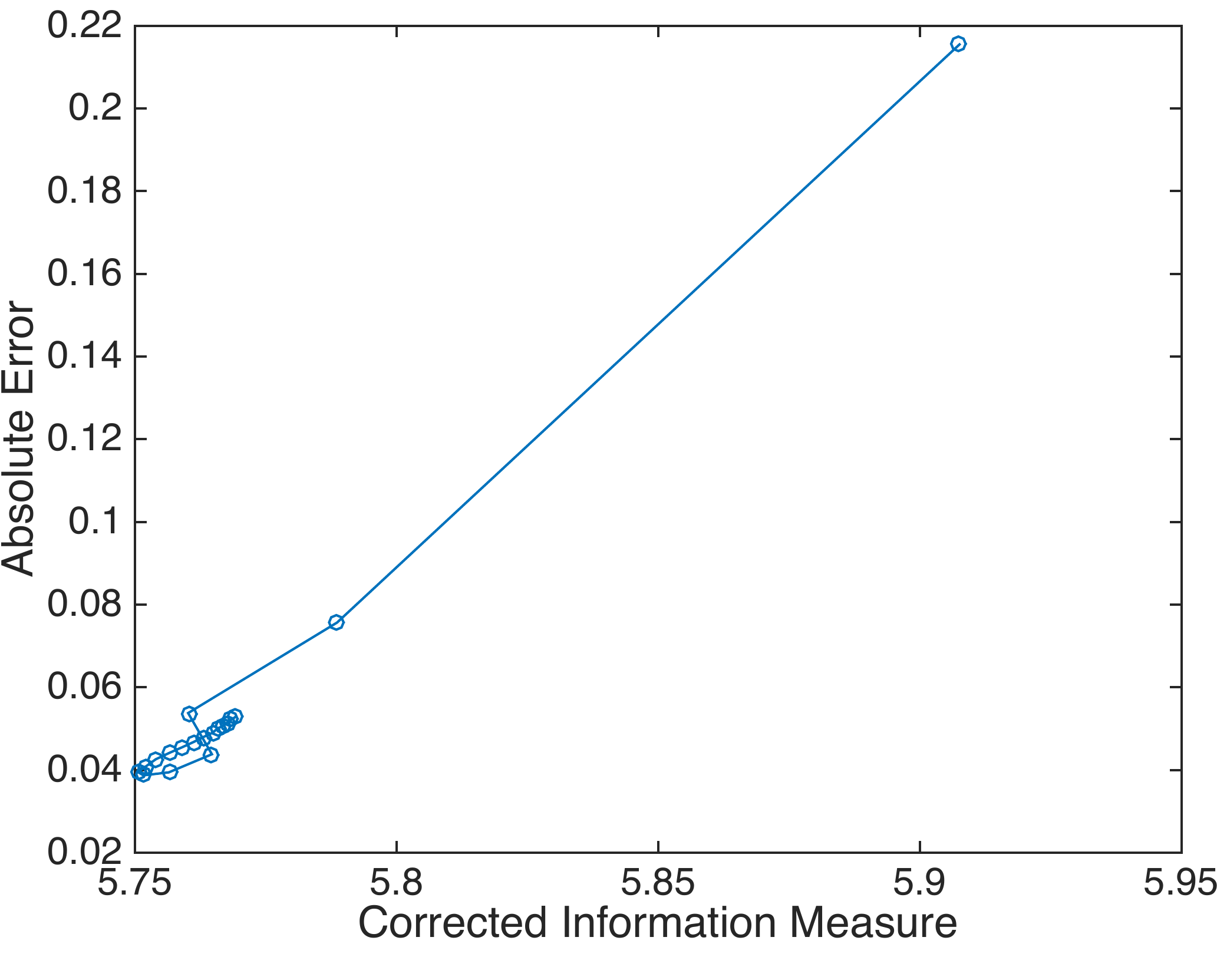}
	\caption{Plot of Absolute Error vs the Corrected information measure for Thermomech TC SuiteSparse dataset on 5 stochastic trace estimate samples}
	\label{fig:cropped_thermomech5samplesgicerror-cropped}
\end{figure}

In the preceding we have assumed that we know the functional expectations of the true eigenvalue distribution. In general we have an estimate, which varies from the true by some error $\epsilon$, i.e.
\begin{equation}
\label{noiseassumption}
\langle x^{i}_{estimate} \rangle = \langle x^{i}_{true}\rangle + \epsilon^{i}.
\end{equation}
Hence Equation \eqref{sumentropies} becomes
\begin{equation}
\label{klwithnoise}
\mathcal{D}(p|q) = -\mathcal{S}(p) +\mathcal{S}(q) + \sum_{i}\alpha_{i}\epsilon^{i}.
\end{equation}
From Equation \eqref{klwithnoise} we can write down a natural measure of constraint informativeness, as
\begin{equation}
\label{iic}
IC = \Delta \mathcal{S}(q) + \sum_{i}\Delta\alpha_{i}\epsilon^{i} \leq 0.
\end{equation}
Where $IC$ is short hand for the informativeness criterion and the latter two terms in Equation \eqref{iic} we hereby refer to as the corrected information measure. We have made explicit the fact that upon the inclusion of an extra constraint all the Lagrange multipliers change in value.
We can approximate the initially intractable \eqref{iic} as:
\begin{equation}
\label{gicvar}
 \Delta \mathcal{S}(q) + \sum_{i}|\Delta\alpha_{i}|\sqrt{Var(\epsilon)} \leq 0.
\end{equation}
In Figure \ref{fig:cropped_thermoent5sampleserrorentropy-croppedy} we plot the entropy of the MaxEnt proposal distribution $\mathcal{S}(q)$ vs absolute error of the Thermomech TC SuiteSparse log determinant for 5 samples. In Figure \ref{fig:cropped_thermomech5samplesgicerror-cropped} we plot the corresponding corrected information measure $\mathcal{S}(q) + \sum_{i}|\alpha_{i}|\sigma_{i}$ against absolute error, where we have used the empirical sample standard deviation as a proxy. We note that the two figures are more or less indistinguishable. We generally find across the SparseSuite dataset, that the corrected information measure provides no discernible benefit in achieving improved performance, even for very small sample numbers. This is due to the inherently low stochastic trace estimate variance for big matrices. We henceforth neglect noise and error in our moment constraints.

\section{KL as a measure of distance}
Using Pinsker's inequality, which is tight up to constant factors, we can relate the KL divergence to both the total variation distance and the total variation norm \cite{cover2012elements}:
\begin{equation}
\delta(P,Q) \leq \sqrt{\frac{1}{2}\mathcal{D}_{kl}(P||Q)},
\end{equation}
where the total variation distance is defined as
\begin{equation}
\delta(P,Q) = sup \{|P(A)-Q(A)|\} \text{where} \in \Sigma. 
\end{equation}
The total variation norm between $P$ and $Q$ can be written as,
\begin{equation}
|P-Q| \leq \sqrt{2\mathcal{D}_{kl}(P||Q)}.
\end{equation}
This follows as $2\delta(P,Q) = |P-Q|_{1}$ where the 1 relates to the $L{1}$ norm.

\subsection{Bound on Error of Log Determinant Approximation}
To calculate the log determinant of the matrix in question, once we have the proposal eigenvalue distribution $q(x)$ we calculate the mean value of $\log(x)$ under the distribution $q(x)$, i.e $\int q(x)\log(x)dx$. We can write the error of our MaxEnt estimate as
\begin{equation}
    \epsilon = \bigg| \int_{x\in \chi}[p(x)-q(x)]\log(x)dx\bigg|
\end{equation}
Where $p(x)$ is the true eigenvalue distribution. $\forall p(x),q(x) \geq 0$ it is hence true that,
\begin{equation}
 \bigg|\int_{x\in \chi}[p(x)-q(x)]\log(x)dx\bigg| \leq  \int_{x\in \chi} |p(x)-q(x)||\log(x)| dx.
\end{equation}
From the monotonicity of the function $\log(x)$, we have that  $\log(x) \leq \mbox{max}[|\log(x_{max})|,|\log(x_{min})|$ and rewriting the total variational norm in terms of the KL divergence we have:
\begin{equation}
\label{errorequation}
\epsilon \leq \mbox{max}[|\log(x_{max})|,|\log(x_{min})|\sqrt{2\mathcal{D}_{kl}(P||Q)}.
\end{equation}
We thus note that by reducing the self entropy of the proposal distribution $q(x)$ we necessarily reduce the maximum possible error of the log determinant estimation. However, given that we do not have an analytic form of $p(x)$, we cannot explicitly calculate $\mathcal{D}_{kl}(P||Q)$ and hence the bound in its current form is not inherently practical. We leave the estimation of this term and derivation of estimate uncertainty to future work.

\section{Algorithm}
We apply a numerically stable MaxEnt Algorithm (algorithm \ref{alg:coefopt}) \cite{bandyopadhyay2005maximum}, under the conditions that $\lambda_i$ is strictly positive and the all power moments $0 \leq \lambda^{k}\leq 1$. 
We can satisfy these conditions by normalizing our positive definite matrix by the maximum of the Gershgorin intervals~\cite{Gershgorin1931}.

\algrenewcommand\algorithmicindent{1.3em}
\renewcommand{\algorithmicrequire}{\textbf{Input:}}
\renewcommand{\algorithmicensure}{\textbf{Output:}}

\begin{algorithm}
\caption{Optimising the Coefficients of the MaxEnt Distribution}\label{alg:coefopt}
\begin{algorithmic}[1]
\vspace{0.5em}
\Require Moments $\{\mu_i\}$, Tolerance $\epsilon$ 
\Ensure Coefficients $\{\alpha_i\}$
\State $\alpha_i \sim \mathcal{N}(0,1)$
\State $i \gets 0$
\State $p(\lambda) \gets \exp(-1 - \sum_k \alpha_k \lambda^k)$
\While{error $< \epsilon$}
\State $\delta \gets \log \left(\frac{\mu_i}{\int \lambda^i p(\lambda) d\lambda} \right)$
\State $\alpha_i \gets \alpha_i + \delta$
\State $p(\lambda) \gets p(\lambda | \alpha)$
\State error $\gets \max |\int \lambda^i p(\lambda) d\lambda - \mu_i|$
\State $i \gets \text{mod}(i+1, \text{length}(\mu))$
\EndWhile 
\end{algorithmic}
\end{algorithm}

We follow the procedure from entropic trace estimation \cite{ete}. Firstly, the raw moments of the eigenvalues are estimated using stochastic trace estimation. These moments are then passed to the maximum entropy optimization of Algorithm \ref{alg:coefopt} to produce an estimate of the distribution of eigenvalues, $p(\lambda)$.
Consequently, $p(\lambda)$ is used to estimate the distribution's log geometric mean, $\int \log(\lambda) p(\lambda) d\lambda$.
This term is multiplied by the matrix's dimensionality and if the matrix was normalized, the log of this normalization term is added. We lay out these steps more concisely in Algorithm \ref{alg:logdet}.  

\algrenewcommand\algorithmicindent{1.3em}
\renewcommand{\algorithmicrequire}{\textbf{Input:}}
\renewcommand{\algorithmicensure}{\textbf{Output:}}

\begin{algorithm}
\caption{Entropic Trace Estimation for Log Determinants}\label{alg:logdet}
\begin{algorithmic}[1]
\vspace{0.5em}
\Require PD Symmetric Matrix $A$, Order of stochastic trace estimation $k$, Tolerance $\epsilon$
\Ensure Log Determinant Approximation $\log|A|$
\State $B = A/\|A\|_2$
\State $\mu$ (moments)$ \gets$ StochasticTraceEstimation$(B, k)$ 
\State $\alpha$ (coefficients) $\gets \text{MaxEntOpt(}\mu, \epsilon)$
\State $p(\lambda) \gets p(\lambda | \alpha)$
\State $\log|A| \gets n\int \log(\lambda) p(\lambda) d\lambda + n\log(\|A\|_2)$
\end{algorithmic}
\end{algorithm}

\section{Algorithmic details for Practitioners}
Given that the MaxEnt approach of Algorithm \ref{alg:coefopt} is numerical, we need to specify a gridding of the input space or choice of nodes. We find that a gridding between $0\leq x \leq 1$ of $\Delta x = 0.001$ provides a good trade-off between speed and accuracy, with essentially the same results (measured by absolute error) as $\Delta x = 0.0001$. We find that the algorithm consistently outputs distributions of increased entropy for more than $m\geq 8$ moment constraints, as is demonstrated in figure \ref{fig:thermomech30sampleerrorvsentropy}, over a variety of data sets. Given that this is independent of the gridding size (tested between $0.01$ and $0.00001$), and given the proof of section \ref{entropydecreaseproof}, showing that an increase in the number of mean value constraints can only decrease the objective value, we consider this indicative of algorithmic break-down and an inability of the algorithm to identify the true global optimum. This explains the increase in relative error with increasing number of moments in \cite{ete} and we hence do not recommend going beyond 8 moments. 

\subsection{Stochastic trace estimates}
\begin{figure}[t]
	\centering 
	\includegraphics[width=0.5\textwidth]{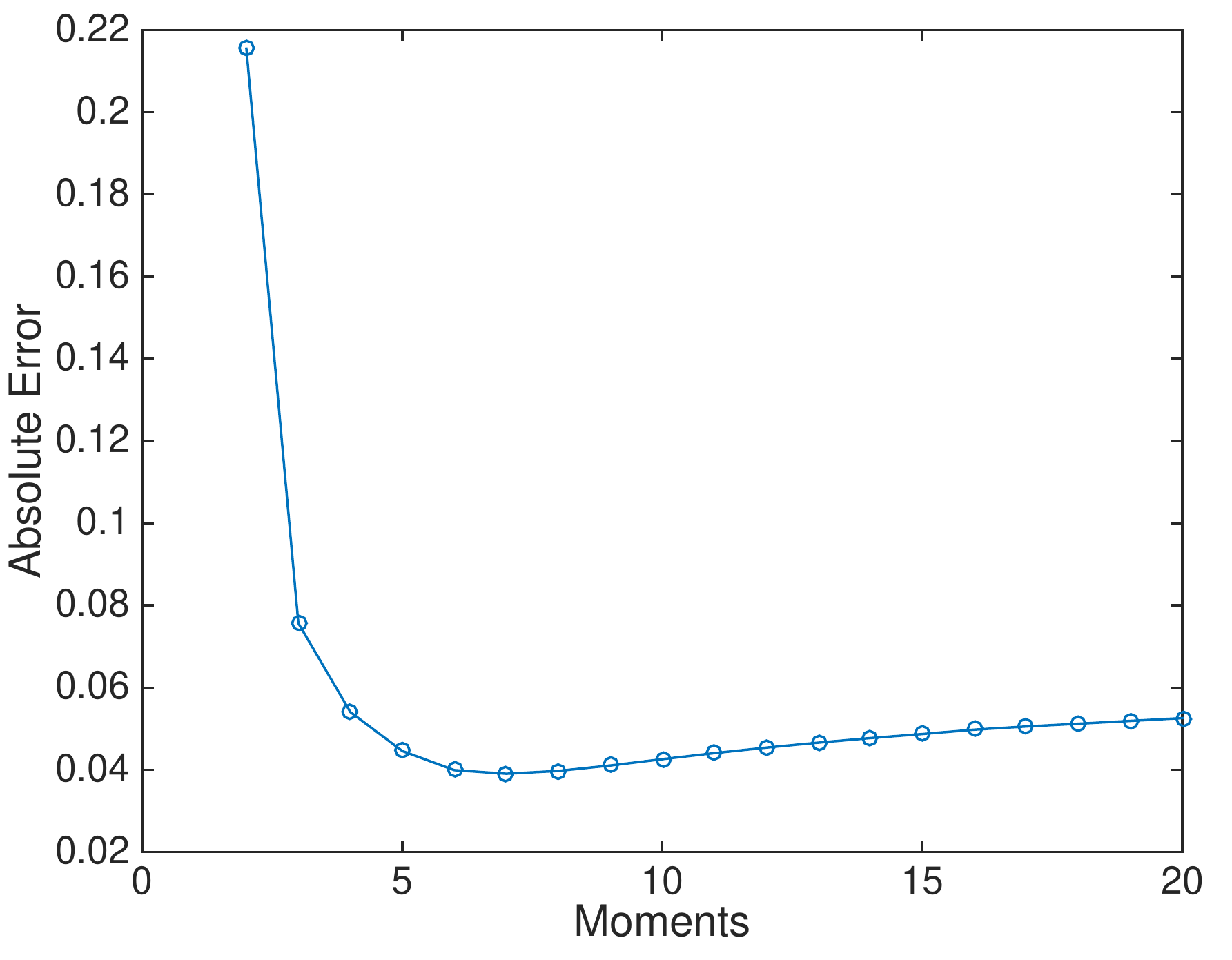}
	\caption{Relative error against number of moments included for a single sample stochastic trace estimate for the Thermomech dataset.}
	\label{fig:thermomech1sampleerrorvsmoments}
\end{figure}

To keep our results comparable and consistent, we keep with \cite{ete,bild} and consider Gaussian random unit vectors. We note that across a variety of sparse datasets, the number of samples taken neither largely effects the entropy of the proposal distribution (used to determine the number of moments required before attaining an optimal result) as is demonstrated by the indiscernability of figures \ref{fig:thermomech1sampleerrorvsentropy} and \ref{fig:thermomech30sampleerrorvsentropy}, nor the absolute error for a given number of moments shown in figures \ref{fig:thermomech1sampleerrorvsmoments} and \ref{fig:thermomech30sampleerrorvsmoments}. Both sets of plots compare a single stochastic trace estimate with an average of 30. We note that the single shot variance of the Gaussian stochastic trace estimator is $2 Tr(A^{2}) $\cite{jackmub}. Intuitively this makes it clear that the standard deviation of a single sample as a fraction of the estimate varies as:
\begin{equation}
\label{asymptoticsamplelaw}
\frac{\sqrt{2 Tr(A^{2})}}{Tr(A)} \approx \sqrt{\frac{\log(n)}{2n}} \xrightarrow{n \rightarrow \infty} 0.
\end{equation}
Here we have used Weyl's inequality that the eigenvalues decay approximately as $\lambda_{n} = \lambda_{max}n^{-1/2}$ \cite{Weyl1912} and that for large matrices we can approximate the sum as an integral, and take the $n\rightarrow \infty$ limit. We thus see that for large matrices we expect the single shot variance to be small and the extra variance reduction by taking more samples may not be required. 

This will depend on the matrix in question and we expect the asymptotic 0 variance result to hold with better accuracy as the size of the matrix increases. Given that the computational cost rises as $\mathcal{O}(n^{2}md)$, where $n$ is the matrix dimension, $m$ is the number of moments and $d$ is the number of samples, we note that for excessively large matrices in which generating the moment estimates takes up a significant portion of run-time. 

\section{Results} 
\subsection{self entropy and absolute error}
\begin{figure}[t]
	\centering 
	\includegraphics[width=0.5\textwidth]{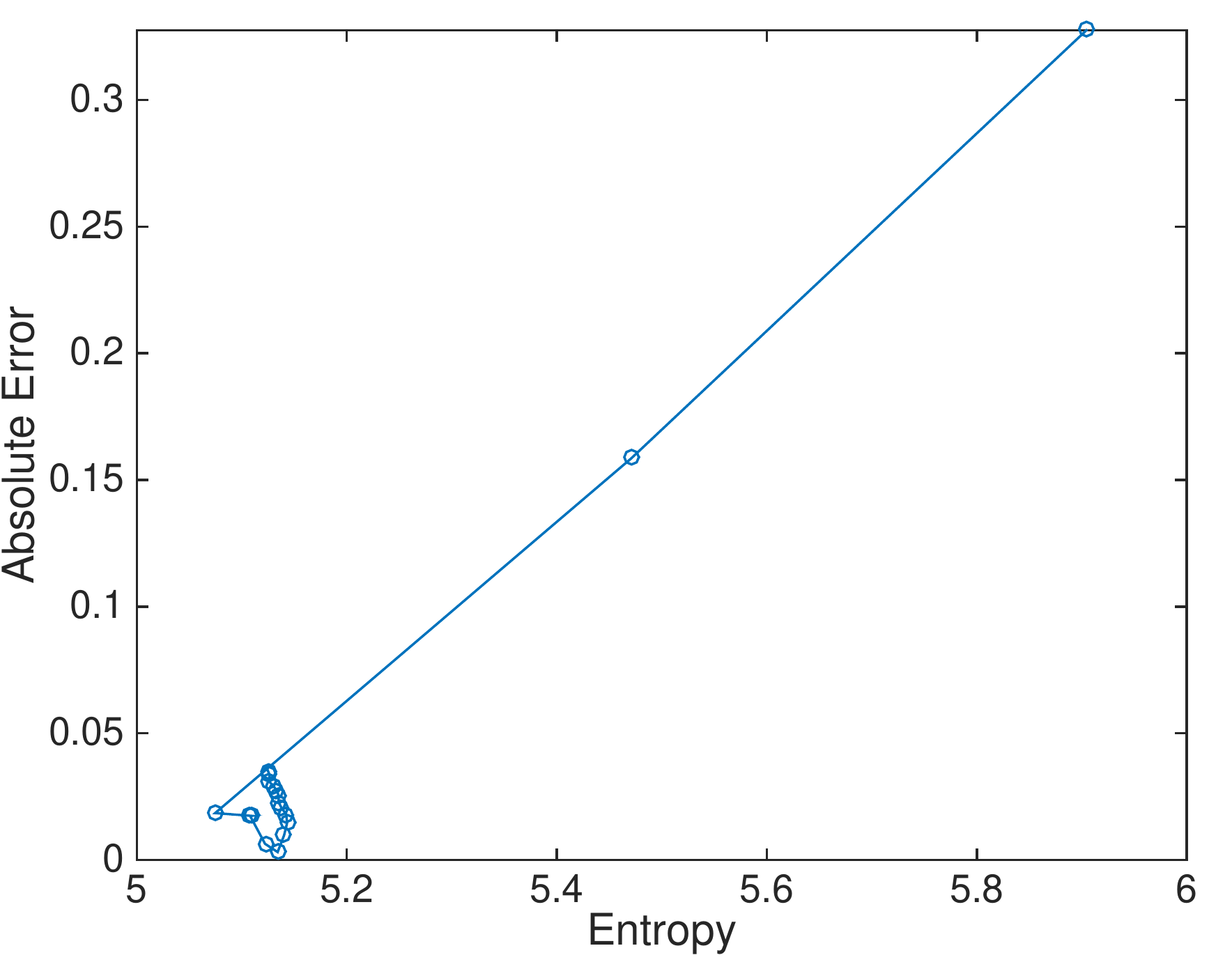}
	\caption{Relative error against entropy for a 30 sample stochastic trace estimate from 2 to 20 moments for the Apache dataset.}
	\label{fig:apacheentropyerror}
\end{figure}

We empirically demonstrate the relationship between proposal distribution self entropy and absolute error on a variety of SuiteSparse datasets, with Figures \ref{fig:thermomech30sampleerrorvsentropy} and \ref{fig:apacheentropyerror} showing the Entropy vs Absolute error for the Thermomech/Apache datasets for $30$ stochastic samples as we increase the number of moment estimates from $2$ to $20$. We see here a general trend of absolute error decreasing as does the self entropy of our proposal distribution, in accordance with our earlier derivations. 

\begin{figure}[t]
	\centering 
	\includegraphics[width=0.5\textwidth]{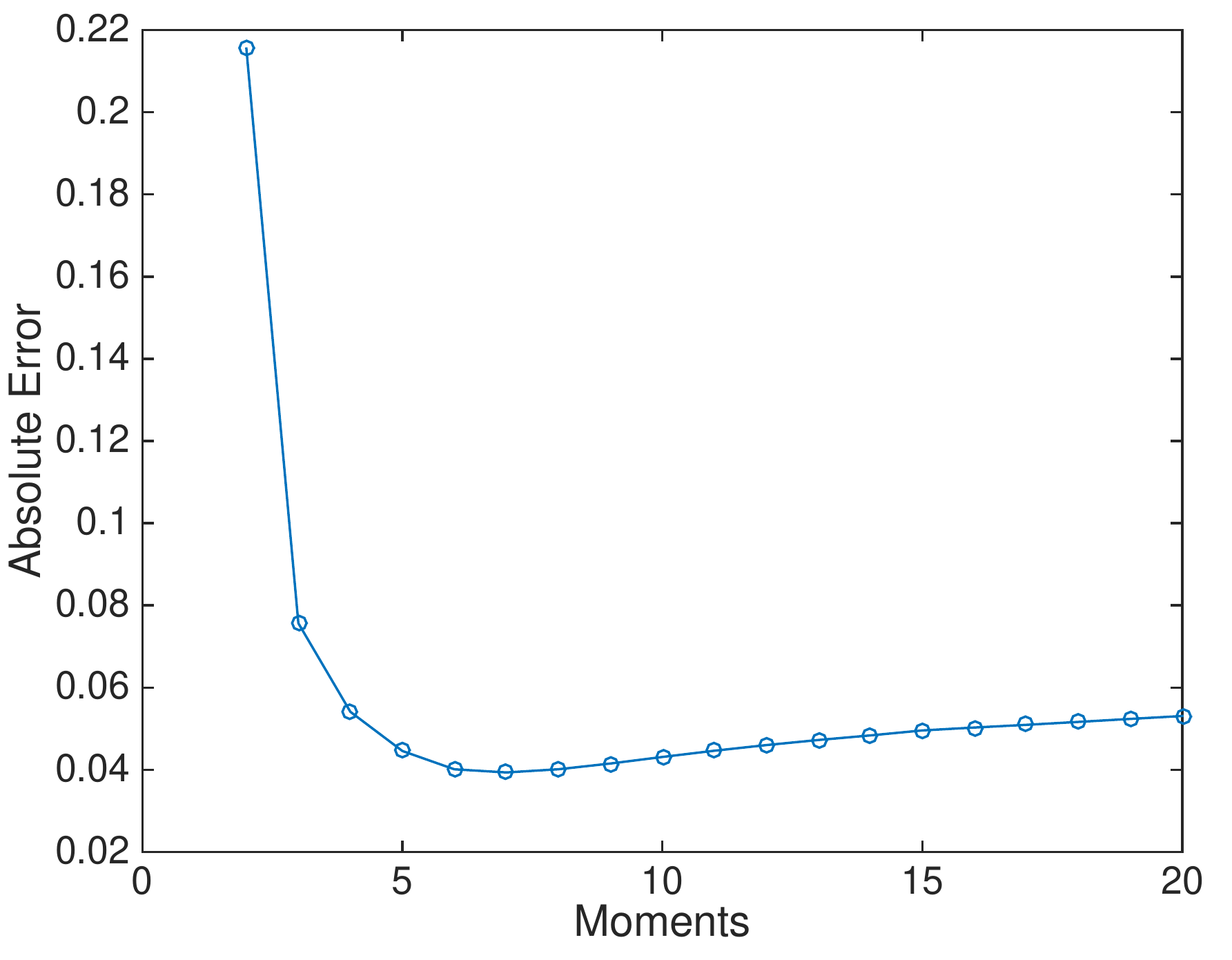}
	\caption{Relative error against number of moments included for a 30 sample stochastic trace estimate for the Thermomech dataset.}
	\label{fig:thermomech30sampleerrorvsmoments}
\end{figure}

\begin{figure}[t]
	\centering 
	\includegraphics[width=0.5\textwidth]{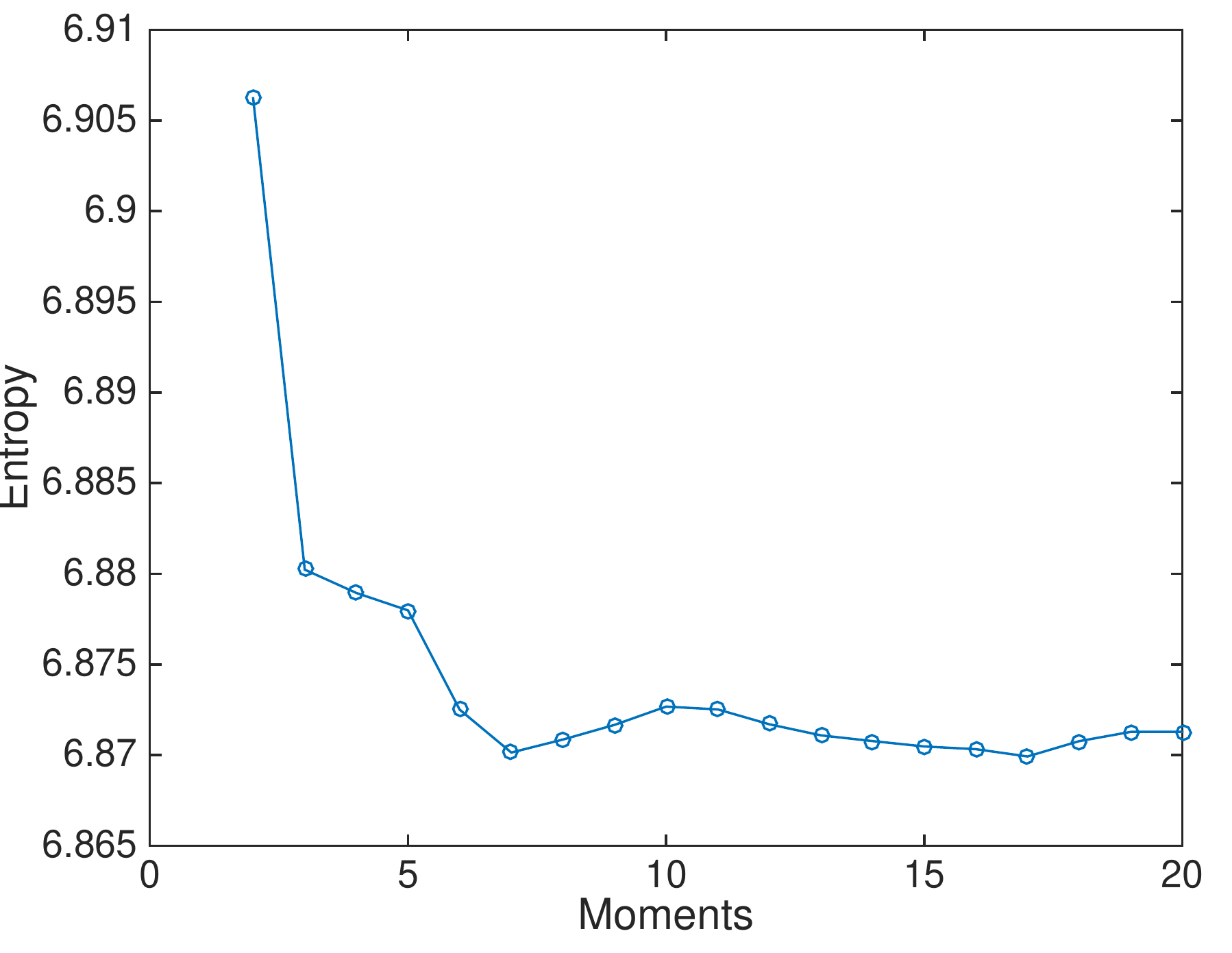}
	\caption{Moments vs Entropy 30 Stochastic Trace samples Ecology2 dataset.}
	\label{fig:momentsentropy_smpl_30_dataset_ecology2}
\end{figure}

\begin{figure}[t]
	\centering 
	\includegraphics[width=0.5\textwidth]{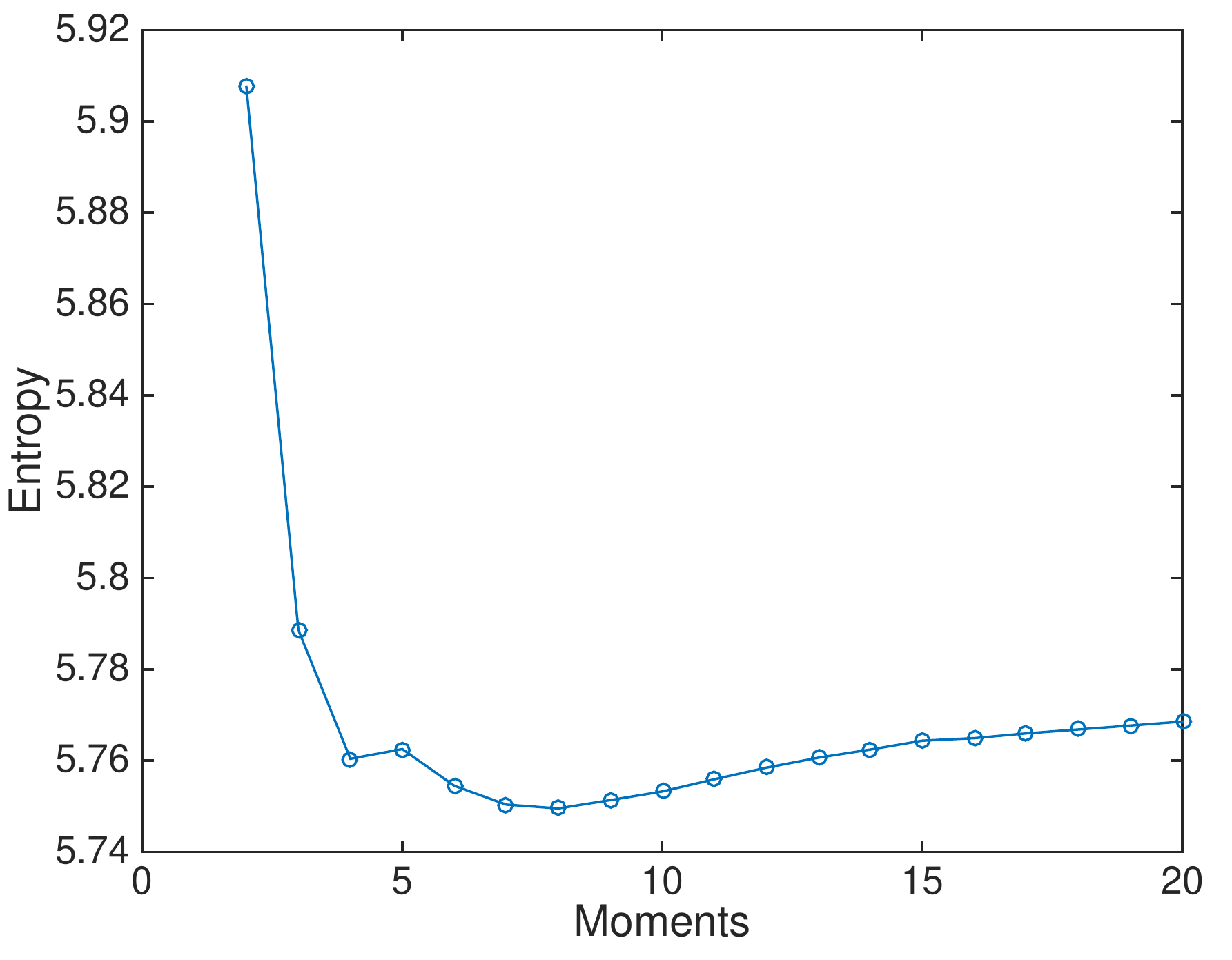}
	\caption{Moments vs Entropy 30 Stochastic Trace samples ThermoMech dataset.}
	\label{fig:momentsentropy_smpl_30_dataset_thermomech_TC}
\end{figure}

We plot the the Entropy vs Moments of the TC and Ecology datasets in Figures \ref{fig:momentsentropy_smpl_30_dataset_ecology2} and \ref{fig:momentsentropy_smpl_30_dataset_thermomech_TC}, where we see a general trend of the entropy reducing monotonically up till a certain number of moments and then increasing or oscillating. We have already proved in section \ref{proofeteworks} that an increase in moments can only decrease the self entropy, which combined with the low sample variance of stochastic trace estimates indicates an inability for the Algorithm \ref{alg:coefopt} to find the true global maximum. The point at which adding extra moments increases the self entropy of the proposal distribution in the algorithm, tends to coincide with a rise in absolute error as we would expect from Equations \eqref{errorequation} and \eqref{kldiv}. We see this experimentally in Figures \ref{fig:thermomech30sampleerrorvsmoments} and \ref{fig:momentsentropy_smpl_30_dataset_thermomech_TC}, where beyond a certain number of moments $m \approx 8$, both the entropy and the absolute relative error begin to rise, this informs our judgment of recommending no more than 8 moments.

\subsection{single sample result comparisons}

We load five sparse (SuiteSparse) square PSD matrices, ranging from a maximum dimension of $999,999$ to a minimum of $81,200$ and run Cholesky using the Matlab 2014b 'Chol' function to calculate the log determinants on a 2.6 GHz Intel Core i7 16 GB 1600 MHz DDR3 notebook. This takes 4847 seconds. Using our MaxEnt Algorithm \ref{alg:coefopt}, with a gridding of $0.001$ and 30 stochastic samples of 8 moments, we calculate the log determinants in 20 seconds. For a single sample, we calculate the log determinants in 16 seconds.  The respective errors are shown in Table \ref{table:experiment 1}.

\begin{table}[h!]
\centering
\begin{tabular}{||c c c c||} 
 \hline
Dataset & Dimension & Samples & Error\\ [0.5ex] 
 \hline\hline
 ecology2 & 999,999 & 30 & 0.0102 \\
  &  & 1 & 0.0105\\
 \hline
 thermomech TC & 102,158 & 30 & 0.0398\\
  & 102,158 & 1 & 0.0402\\
 \hline
 shallow water 1 & 81,920 & 30 & 0.0043\\
 &  & 1 & 0.0035\\
 \hline
 shallow water 2 & 81,920 & 30 & 0.0039\\
 &  & 1 & 0.0040\\
 \hline
 apache1 & 80,800 & 30 & 0.006571\\
 &  & 1 & 0.0101\\ [0.5ex] 
 \hline
\end{tabular}
\newline
\caption{Relative absolute error on SuiteSparse datasets, for 8 moments and 30/1 samples per moment}
\label{table:experiment 1}
\end{table}
We note that even for relatively small matrices, $n \approx 80,000$, that the performance from reducing the number of samples is relatively unaffected. This suggests that the asmyptotics from Equation \eqref{asymptoticsamplelaw} come into play rather early on. However, given that there is a slight decrease in performance and that 15 of the 20 and 16 seconds of compute time were spent on Alogorithm \ref{alg:coefopt}, determining the Maximum Entropy coefficients, we recommend reducing the number of samples only when it becomes a larger proportion of the overall cost. 

We further test the validity and practical value of the heuristic derivation of equation \eqref{asymptoticsamplelaw}, by evaluating the difference in MaxEnt estimate for the largest PSD matrices in the SuiteSparse data set, comparing a single sample to $30$ samples. For $Queen 4147$ with a dimension of $4,147,110$ and $316,548,962$ non zero values, the difference in prediction from taking $1$ sample instead of $30$ is $0.0028\%$ and the run-time is reduced from $173$ to $60$ seconds. We note that the standard Cholesky and LU functions in (e.g.) MATLAB are unable to handle matrices of that size, due to contiguous memory constraints, even on significantly more powerful machines than the one above. Table \ref{table:experiment 2} shows results for a variety of large datasets.
\begin{table}[h!]
\centering
\begin{center}
 \begin{tabular}{||c c c c c c||} 
 \hline
Dataset & Dimension & Samples & Estimate & Time(s)  & \textbf{$\Delta \%$} \\ [0.5ex] 
 \hline\hline
 Queen & 4,147,110 & 30 & -7.3951e+07 & 172.4 & \\
 Non 0's & 316,548,962 & 1 & -7.3953e+07 & 60.3 & 0.0028 \\ 
 \hline
 Bump & 2,911,419 & 30 & -5.2282e+07 & 64.3 &\\ 
 Non 0's & 127,729,899 & 1 & -5.2297e+07 & 15.9 & 0.029\\ 
 \hline
Serena & 1,391,349 & 30 & -1.5831e+07 & 34.4 & \\ 
Non 0's & 64,131,971 & 1 & -1.5771e+07 & 8.867 & 0.38\\ 
 \hline
Geo & 1,437,960 & 30 & -1.0186e+07 & 33.2 &\\ 
Non 0's & 60,236,322 & 1 & -1.0203e+07 & 12.5 & 0.17\\
 \hline
Hook & 1,498,023 & 30 & -4.6026e+06 & 32.3 & \\  
Non 0's & 59,374,451 & 1 & -4.6033e+06 & 11.8 & 0.015\\
 \hline
StochF & 1,465,137 & 30 & -2.6807e+07 & 15.4 & \\ 
Non 0's & 21,005,389 & 1 & -2.6812e+07 & 7.1 & 0.019\\
 \hline
G3 & 1,585,478 & 30 & -1.0263e+07 & 9.875 & \\  
Non 0's & 7,660,826 & 1 & -1.0262e+07 & 4.618 & 0.097 \\ [0.5ex] 
 \hline
\end{tabular}
\end{center}
\caption{results for the largest psd suitesparse matrices, using 8 moments, with sample number either 30 or 1. Final column denotes percentage difference in estimate between using 30/1 sample(s).}
\label{table:experiment 2}
\end{table}
The reduction in samples from $30$ to $1$ reduces the computational run-time by a factor of $3$ and the difference in estimates, which is always less than $0.4\%$ tends to increase in as the matrix dimension decreases, in accordance to our asymptotic law of Equation \eqref{asymptoticsamplelaw}. The exceptions, $StochF$ and $G3$, are both significantly sparser than the others, which is why they run significantly faster and the MaxEnt calculation algorithm (which is independent of the number of samples taken) takes up a greater proportion of the total run-time and hence the reduction from taking less samples is less. We posit a potential link between sparsity and accuracy, but leave the investigation for future work.

\begin{figure}[t]
	\centering 
	\includegraphics[width=0.5\textwidth]{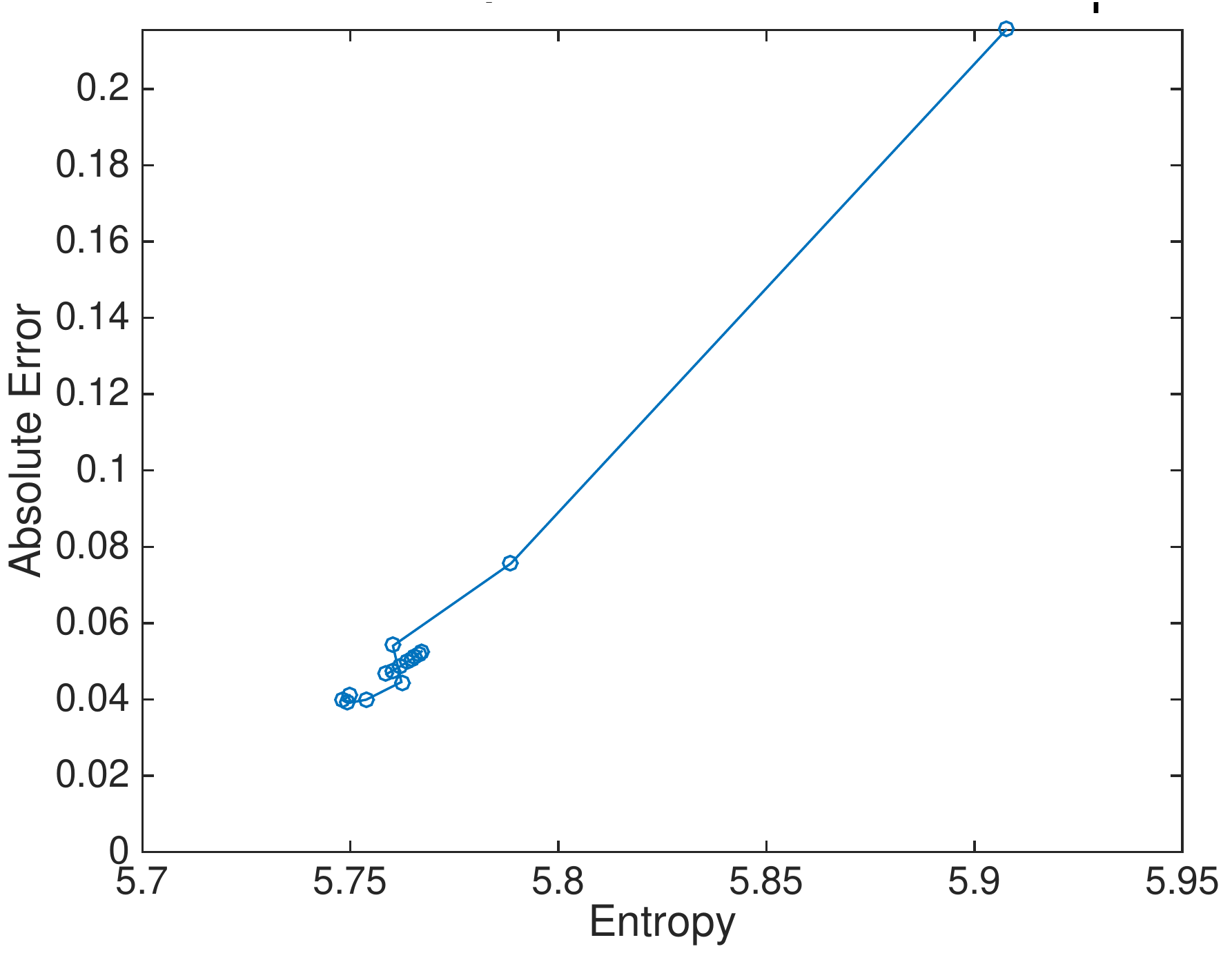}
	\caption{Relative error against entropy for a single sample stochastic trace estimate from 2 to 20 moments for the Thermomech dataset.}
	\label{fig:thermomech1sampleerrorvsentropy}
\end{figure}

\begin{figure}[t]
	\centering 
	\includegraphics[width=0.5\textwidth]{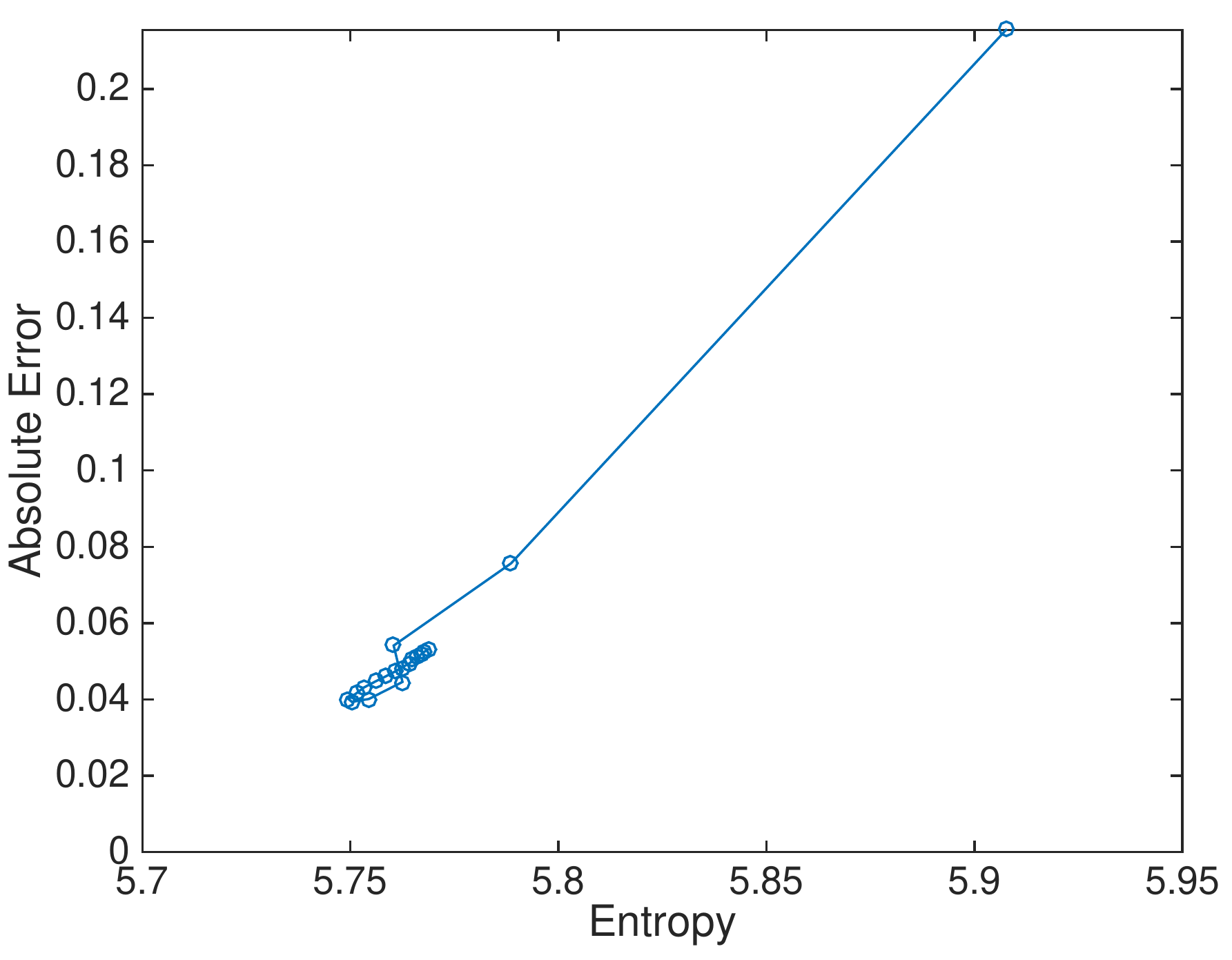}
	\caption{Relative error against entropy for a 30 sample stochastic trace estimate from 2 to 20 moments for the Thermomech dataset.}
	\label{fig:thermomech30sampleerrorvsentropy}
\end{figure}

The link between reduction in proposal self entropy and absolute error, is also unchanged as we reduce the number of samples, as can be seen by comparing Figures \ref{fig:thermomech1sampleerrorvsentropy} and \ref{fig:thermomech30sampleerrorvsentropy}, if the variance of samples were higher this would not be the case, as Equation \eqref{iic} would not be $\leq 0$ due to the error term and we would not have a corresponding reduction in KL divergence or $L1$ norm.


\section{Conclusion}
In this paper we formally establish the link between sample expectation and mean value constraint, proving asymptotic equivalence. We also prove that the eigenvalue distribution of a Covariance matrix can be uniquely determined by its moments. The combination of these two provides a solid foundation for using stochastic trace estimation sample estimates as mean value constraints for a Maximum Entropy estimation of a Covariance matrix eigenvalue density.

We further show how the inclusion of extra moment constraints, necessarily reduces the KL divergence $\mathbb{D}_{kl}(p||q)$ between the MaxEnt proposal $q$ and true eigenvalue distribution $p$ and relate this to the $L1$ norm, deriving a novel error bound on Log Determinants using MaxEnt. We demonstrate empirically on SuiteSprase datasets how this reduction in $\mathbb{D}_{kl}(p||q)$ corresponds to increased estimation accuracy. 

We investigate the effect of reducing the number of stochastic trace estimate samples empirically and provide an informal derivation of an asymptotic law, which we verify experimentally; the larger the matrix, the smaller the effect and the greater the computational benefit of reducing the number of samples. 

We set up best practice guidelines, rooted in theory and experiment, for practitioners wishing to deal with large matrices. Our basic, non-optimized MaxEnt implementation is able to calculate determinants of $4$ million by $4$ million matrices on a laptop within a minute.

\section{Future Work}

Future work could involve the investigation of the inability of the current algorithms inability to find a true global optimum beyond a certain number of moments and potentially remedy this by using Chebyshev polynomials or another orthogonal polynomial set in the optimisation. The estimation of the $\mathcal{D}_{kl}(p||q)$ to be able to practically use the derived bound and the extension of the above to Matrix inversion.


%

\section*{Acknowledgment}
The authors would like to thank the Oxford Man Institute for their grant in supporting this research, Thomas Gunter and Michael Osborne for illuminating discussions, Pawan Kumar for his input on Convex Analysis and Tim Davies for the upkeep or the SuiteSparse dataset.

\ifCLASSOPTIONcaptionsoff
  \newpage
\fi

\bibliographystyle{unsrt}
	\bibliography{sample}

\begin{thebibliography}{10}

\bibitem{Bai1997}
Zhaojun Bai and Gene~H. Golub.
\newblock Bounds for the {T}race of the {I}nverse and the {D}eterminant of
  {S}ymmetric {P}ositive {D}efinite {M}atrices.
\newblock {\em Annals of Numerical Mathematics}, 4:29--38, 1997.

\bibitem{Golub1996}
Gene~H. Golub and Charles~F. Van~Loan.
\newblock {\em {Matrix computations}}.
\newblock The Johns Hopkins University Press, 3rd edition, October 1996.

\bibitem{Macchi1975}
Odile Macchi.
\newblock The {C}oincidence {A}pproach to {S}tochastic point processes.
\newblock {\em Advances in Applied Probability}, 7:83--122, 1975.

\bibitem{Rasmussen2006}
Carl~E. Rasmussen and Christopher Williams.
\newblock {\em {Gaussian Processes for Machine Learning}}.
\newblock MIT Press, 2006.

\bibitem{Wainwright2006}
Martin~J. Wainwright and Michael~I. Jordan.
\newblock Log-determinant relaxation for approximate inference in discrete
  markov random fields.
\newblock {\em {IEEE} Trans. Signal Processing}, 54(6-1):2099--2109, 2006.

\bibitem{ete}
Jack Fitzsimons, Diego Granziol, Kurt Cutajar, Michael Osborne, Maurizio
  Filippone, and Stephen Roberts.
\newblock Entropic trace estimates for log determinants, 2017.

\bibitem{billingsley_2012}
Patrick Billingsley.
\newblock {\em Probability and measure}.
\newblock Wiley, 2012.

\bibitem{maxentreview}
Steve Press\'e, Kingshuk Ghosh, Julian Lee, and Ken~A. Dill.
\newblock Principles of maximum entropy and maximum caliber in statistical
  physics.
\newblock {\em Rev. Mod. Phys.}, 85:1115--1141, Jul 2013.

\bibitem{inftheoryjaynes}
E.~T. Jaynes.
\newblock Information theory and statistical mechanics.
\newblock {\em Phys. Rev.}, 106:620--630, May 1957.

\bibitem{giffin}
Adom Giffin, Carlo Cafaro, and Sean~Alan Ali.
\newblock Application of the maximum relative entropy method to the physics of
  ferromagnetic materials.
\newblock {\em Physica A: Statistical Mechanics and its Applications}, 455:11
  -- 26, 2016.

\bibitem{nerioptions}
Cassio Neri and Lorenz Schneider.
\newblock Maximum entropy distributions inferred from option portfolios on an
  asset.
\newblock {\em Finance and Stochastics}, 16(2):293--318, 2012.

\bibitem{entropybuchen}
Peter~W Buchen and Michael Kelly.
\newblock The maximum entropy distribution of an asset inferred from option
  prices.
\newblock {\em Journal of Financial and Quantitative Analysis},
  31(01):143--159, 1996.

\bibitem{diego}
Diego Granziol and Stephen Roberts.
\newblock An information and field theoretic approach to the grand canonical
  ensemble, 2017.

\bibitem{Gonzalez2014}
Diego Gonz{\'a}lez, Sergio Davis, and Gonzalo Guti{\'e}rrez.
\newblock Newtonian dynamics from the principle of maximum caliber.
\newblock {\em Foundations of Physics}, 44(9):923--931, 2014.

\bibitem{caticha2012entropic}
A~Caticha.
\newblock Entropic inference and the foundations of physics (monograph
  commissioned by the 11th brazilian meeting on {B}ayesian
  statistics--ebeb-2012, 2012.

\bibitem{shore1980axiomatic}
John Shore and Rodney Johnson.
\newblock Axiomatic derivation of the principle of maximum entropy and the
  principle of minimum cross-entropy.
\newblock {\em IEEE Transactions on information theory}, 26(1):26--37, 1980.

\bibitem{stein1972}
Charles Stein.
\newblock A bound for the error in the normal approximation to the distribution
  of a sum of dependent random variables.
\newblock In {\em Proceedings of the Sixth Berkeley Symposium on Mathematical
  Statistics and Probability, Volume 2: Probability Theory}, pages 583--602,
  Berkeley, Calif., 1972. University of California Press.

\bibitem{convergenceclt}
A.~N. Tikhomirov.
\newblock On the convergence rate in the central limit theorem for weakly
  dependent random variables.
\newblock {\em Theory of Probability 'l\&' Its Applications}, 25(4):790–809,
  1981.

\bibitem{cover2012elements}
Thomas~M Cover and Joy~A Thomas.
\newblock {\em Elements of information theory}.
\newblock John Wiley \& Sons, 2012.

\bibitem{jaynes1982rationale}
Edwin~T Jaynes.
\newblock On the rationale of maximum-entropy methods.
\newblock {\em Proceedings of the IEEE}, 70(9):939--952, 1982.

\bibitem{boyd_vandenberghe_2009}
Stephen~P. Boyd and Lieven Vandenberghe.
\newblock {\em Convex optimization}.
\newblock Cambridge University Press, 2009.

\bibitem{walley_1991}
Peter Walley.
\newblock {\em Statistical reasoning with imprecise probabilities}.
\newblock Chapman and Hall, 1991.

\bibitem{bandyopadhyay2005maximum}
K~Bandyopadhyay, Arun~K Bhattacharya, Parthapratim Biswas, and DA~Drabold.
\newblock Maximum entropy and the problem of moments: A stable algorithm.
\newblock {\em Physical Review E}, 71(5):057701, 2005.

\bibitem{Gershgorin1931}
Semyon Gershgorin.
\newblock Uber die {A}bgrenzung der {E}igenwerte einer {M}atrix.
\newblock {\em Izvestija {A}kademii {N}auk {SSSR}, {S}erija {M}atematika},
  7(3):749--754, 1931.

\bibitem{bild}
Jack Fitzsimons, Kurt Cutajar, Michael Osborne, Stephen Roberts, and Maurizio
  Filippone.
\newblock Bayesian inference of log determinants, 2017.

\bibitem{jackmub}
J.~K. Fitzsimons, M.~A. Osborne, S.~J. Roberts, and J.~F. Fitzsimons.
\newblock Improved stochastic trace estimation using mutually unbiased bases,
  2016.

\bibitem{Weyl1912}
Hermann Weyl.
\newblock Das asymptotische verteilungsgesetz der eigenwerte linearer
  partieller differentialgleichungen (mit einer anwendung auf die theorie der
  hohlraumstrahlung).
\newblock {\em Mathematische Annalen}, 71(4):441--479, Dec 1912.

\end{thebibliography}
\end{document}